\documentclass[]{spie} 

\usepackage{amsmath,amsfonts,amssymb}
\usepackage{graphicx}
\usepackage[colorlinks=true, allcolors=blue]{hyperref}
\usepackage{algorithm}
\usepackage{algorithmic}
\usepackage{todonotes}
\usepackage{caption}
\usepackage{subfig}
\usepackage{amsmath,amsfonts,amssymb}
\usepackage{graphicx}
\usepackage{float}
\usepackage{xcolor}
\usepackage{ulem}

\newcommand{\figcaption}[1]{\def\@captype{figure}\caption{#1}}
\newcommand{\tblcaption}[1]{\def\@captype{table}\caption{#1}}

\newcommand{\printfnsymbol}[1]{%
  \textsuperscript{\@fnsymbol{#1}}%
}
\renewcommand\footnotemark{}

\title{Automatic Paper Summary Generation \\from Visual and Textual Information}
\author[*1]{Shintaro Yamamoto\thanks{* First two authors contributed equally.}}
\author[*1]{Yoshihiro Fukuhara}
\author[2]{Ryota Suzuki}
\author[3]{\\Shigeo Morishima}
\author[2]{Hirokatsu Kataoka}

\affil[1]{Waseda University}
\affil[2]{National Institute of Advanced Industrial Science and Technology (AIST)}
\affil[3]{Waseda Research Institute for Science and Engineering}

\begin{document}
\maketitle

\begin{abstract}
Due to the recent boom in artificial intelligence (AI) research, including computer vision (CV), it has become impossible for researchers in these fields to keep up with the exponentially increasing number of manuscripts.
In response to this situation, this paper proposes the paper summary generation (PSG) task using a simple but effective method to automatically generate an academic paper summary from raw PDF data.
We realized PSG by combination of vision-based supervised components detector and language-based unsupervised important sentence extractor, which is applicable for a trained format of manuscripts.
We show the quantitative evaluation of ability of simple vision-based components extraction, and the qualitative evaluation that our system can extract both visual item and sentence that are helpful for understanding.
After processing via our PSG, the 979 manuscripts accepted by the Conference on Computer Vision and Pattern Recognition (CVPR) 2018 are available\footnote[2]{\url{https://cvpaperchallenge.github.io/AutoPaperSummaryGen/}}.
It is believed that the proposed method will provide a better way for researchers to stay caught with important academic papers.

\end{abstract}


\section{Introduction}
\begin{figure}[t]
\begin{center}
\centering
\includegraphics[width=0.45\textwidth]{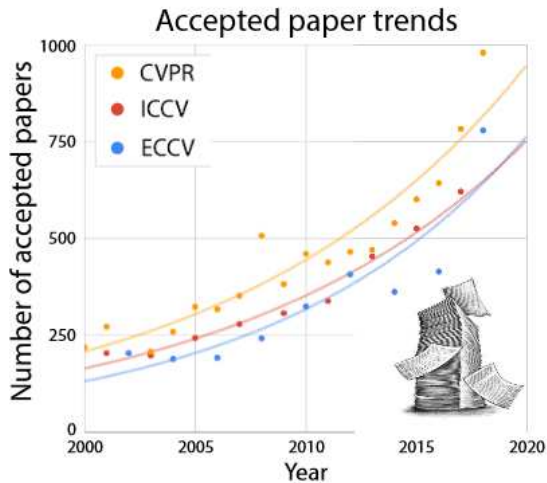}
\figcaption{Trends in the number of papers accepted by top-tier conferences.}
\label{fig:trends}
\end{center}
\end{figure}

Reading papers accepted by top-tier conferences is one of the most effective ways for researchers to stay on top of the trends taking place in their entire field as well as the latest research results.
However, the number of top-tier conference papers in computer vision (CV) has increased tremendously and has effectively tripled in the past 10 years (Fig.~\ref{fig:trends}).
In 2017, more than 1,400 papers were accepted by CVPR and ICCV, and it would require about 700 hours (around 90 days of reading 8 hours per day) to read them all, assuming each paper could be read in 30 minutes.
Unless this trend slows or stops, researchers hoping to stay fully informed in their fields will be required to spend more than 180 days reading papers in 2020, even when using reading for the entire work day.
In addition, due to the fact that the number of notable papers now being published in open access repositories such as arXiv is increasing day by-day, it is not an exaggeration to say that the situation becoming intolerable.
Furthermore, the rapid increase in submissions has resulted in an excess of 3,300 manuscripts for the Conference on Computer Vision and Pattern Recognition (CVPR) 2018, which is imposing an extremely heavy burden on the reviewers, as 2,385 excellent researchers were engaged as reviewers \cite{cvpr2018}.
Such situations hinder the activities of excellent researchers and even degrade review quality.
To make matters worse, some peer reviewers may die while trying to keep pace with the exponentially increasing number of submitted papers in the future, which is an unpleasant scenario that must be avoided if we are to save our community and ensure the continuing progress of humans.
Although the paper entitled \textit{Paper Gestalt} \cite{von2010paper} was written in 2010 as a humorous way to call attention to the problem of even-increasing manuscript submissions, the topic itself is no longer a laughing matter.

To address the problems associated with the explosively increasing papers in the CV community, we propose a novel vision and language task called paper summary generation (PSG), which aims an automatic summary generation from PDF file of an academic paper.
The paper summary must be effective in order to reduce the burden of the reader's understanding because excellent summaries can of course help readers to comprehend enormous papers by providing preliminary knowledge.
For reviewers, an efficient reviewing process is facilitated by understanding the framework before reading the entire paper, which is also achieved by the excellent summary.
As a result, the reader or the reviewer can concentrate on understanding difficult concepts such as mathematical formulas or novelty and validity by reading the entire paper, which is impossible task for artificial intelligence.
In addition, an automatically generated summary can be useful for authors who wish to provide summaries that assist reader understanding, as was required by CVPR until 2015.
For the above reason, we believe that PSG solves the paper bottleneck problem by assisting the understanding of reader or reviewer.

\begin{figure}[t]
\begin{center}
\centering
\includegraphics[width=\linewidth]{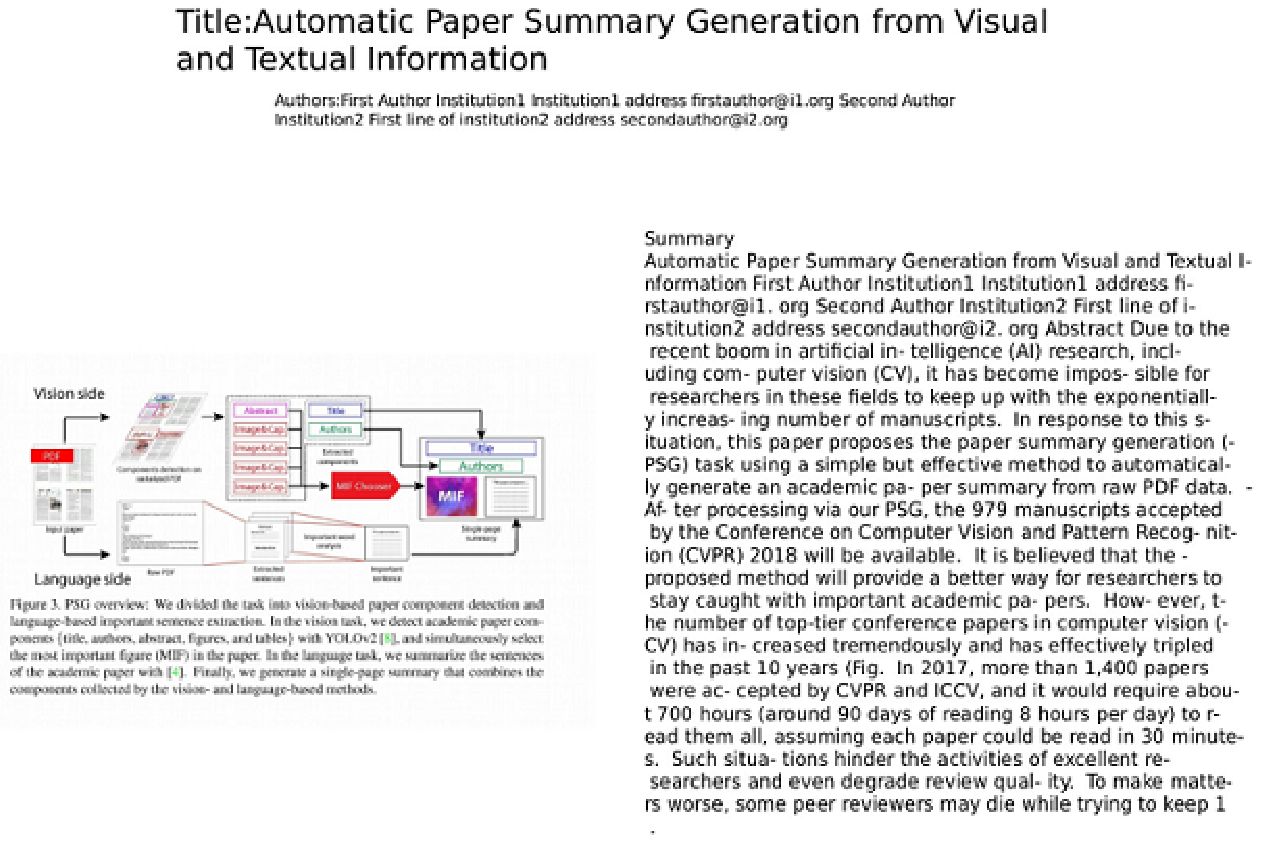}
\figcaption{Sample result of the proposed method. The input is this paper itself.}
\label{fig:sumary_this_paper}
\end{center}
\end{figure}

The main contributions of the present paper are as follows:
(1) We propose a combined vision and language method to automatically generate a single-page summary from a paper as a PSG baseline.
(2) We will release the summaries of the 979 papers accepted by CVPR 2018 that were generated by the proposed method.
In order to reduce demands on the valuable time of reviewers, a summary of this paper that was generated by our proposed method is shown in Fig~\ref{fig:sumary_this_paper}.


\section{Related Work}
\label{sec:related}
Reading a sophisticated summary is one of the most efficient ways to gain an overall understanding of long and difficult sentences.
Generating a summary while retaining the entire context and the important information extracted from a given text has widely been researched in the natural language processing (NLP) community following the remarkable work by Luhn \cite{luhn1958automatic} in 1958.
In recent years, citation-based summarization methods \cite{mei2008generating,qazvinian2008scientific,abu2011coherent} have succeeded in facilitating the academic paper summarization task.
Those methods assume that papers citing the target paper have some parts that briefly mention important aspects of the target paper.
However, we believe that not only linguistic information (such as text) but also visually meaningful items (such as figures and tables) are essential for effective understanding.
In the following paragraphs, we will introduce related research on our automatic summarization method, which considers both items.

Poster presentations, which are considered to be one type of summary, are made by extracting particularly important parts from original papers.
In their work, Qiang et al. \cite{qiang2017learning} proposed a method that can be used to automatically generate an aesthetic poster with high legibility using sentences and figure extracted from a paper.
Hiraoka et al.~\cite{hiraoka2018importance} proposed a method that can be used to automatically estimate the importance of each figure in a paper by determining the importance of sentences citing those figures. However, both these methods require user interaction to extract the figures from the paper.

\section{Proposed Method}
\label{sec:method}
\begin{figure}[t]
\begin{center}
\centering
\includegraphics[width=\linewidth]{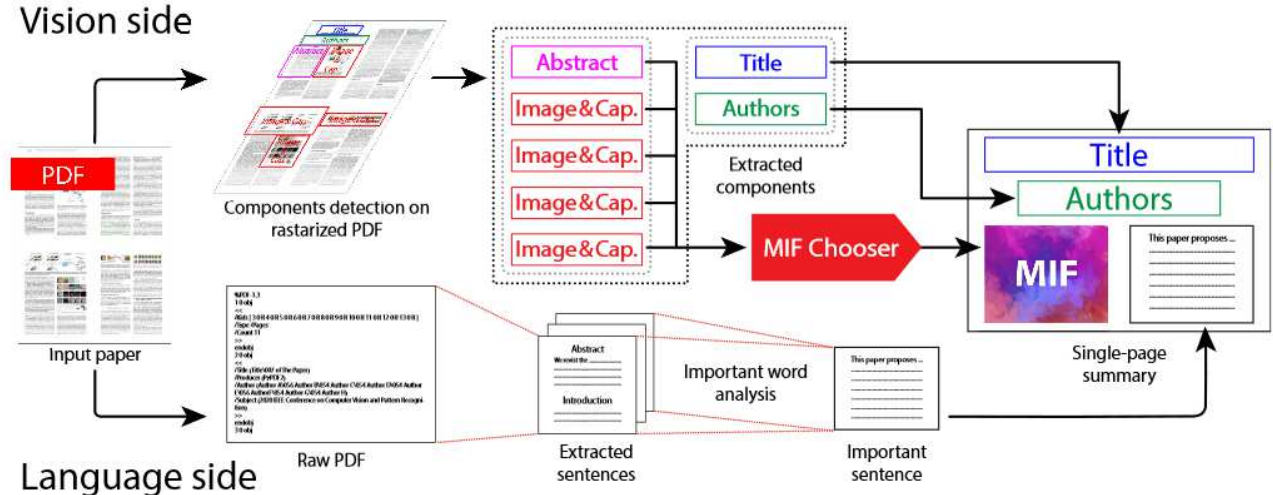}
\caption{PSG overview: We divided the task into vision-based paper component detection and language-based important sentence extraction. In the vision task, we detect academic paper components \{title, authors, abstract, figures, and tables\} with YOLOv2~\cite{redmon2017yolo9000}, and simultaneously select the most important figure (MIF) in the paper. In the language task, we summarize the sentences of the academic paper with~\cite{luhn1958automatic}. Finally, we generate a single-page summary that combines the components collected by the vision- and language-based methods.}
\label{fig:overview}
\end{center}
\end{figure}

In this section, we introduce the PSG method, which is simple but effective to summarize an academic paper from PDF file.
We aim to generate a single-page summary, as the reader does not need to turn the page and it is clear at a glance.
Fig.~\ref{fig:overview} shows an overview of our PSG method.
The system is constructed using both vision-based component extraction and language-based sentence summarization.
The summarized paper includes the paper title, authors, figure, (Sec.~\ref{sec:visual} and \ref{sec:MIF}) and summary sentences (Sec.~\ref{sec:sentence}).

\subsection{Appearance-based paper components extraction}
\label{sec:visual}
To extract the contents of an academic paper, visual information is helpful.
For example, the title, authors and abstract of a paper can readily be identified based on their format.
Thus, we consider the title, authors and abstract to be kinds of information that are  detectable using visual information.
Visual information is essential in other ways, as it can convey the main ideas presented in the paper such as the method overview and experimental results.

\begin{figure}[t]
\begin{center}
\centering
\includegraphics[width=.55\linewidth]{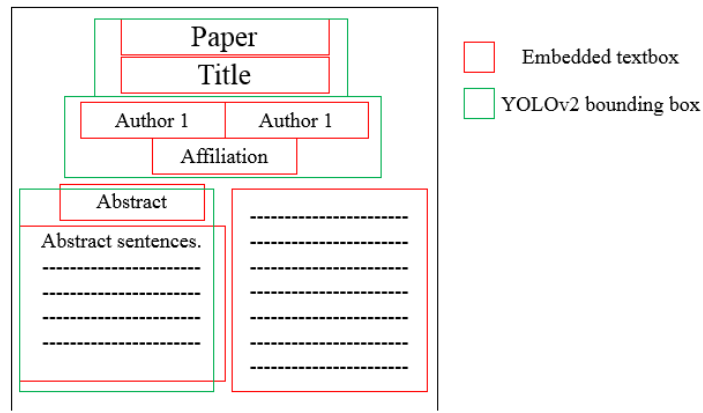}
\figcaption{Text boxes embedded in PDF file. Those corresponding to the abstract, author, and title are divided into a plurality.}
\label{fig:textbox}
\end{center}
\end{figure}

We employ the state of the art YOLOv2~\cite{redmon2017yolo9000} object detection method to obtain visually relevant characteristics (title, author and abstract)\footnote{These contents may be readily detectable with the language-based method, but can also be easily detected with the vision-based since academic papers have fixed layouts.} and visually meaningful (figure and table) contents.
Specifically, we obtain bounding boxes for five object type (abstract, author, figures, tables, and title) from the captured images of each PDF page, and then extract the areas that show the highest confidence level as being the abstract, author, and title from the first page.
We also extract figures and tables that can be detected at high confidence levels.
Based on YOLOv2 bounding boxes, we estimate the text boxes corresponding to the title, authors, and, abstract embedded in PDF file.
Text boxes are divided into a plurality for each item, as shown in Fig.~\ref{fig:textbox}, then we extract ones at high recall.
The definition of recall is as follows:
\begin{equation}
recall=\frac{\mid R_{YOLO} \cap R_{Text}\mid}{\mid R_{Text} \mid}
\end{equation}
Here, $R_{YOLO}$ and $R_{Text}$ are regions indicated by bounding box detected by YOLOv2 and embedded text box in PDF file, respectively. $\mid R \mid$ is defined as the number of pixels inside Region R.

\subsection{Most informative figure}
\label{sec:MIF}
Figures in an academic paper convey a variety of information from auxiliary explanations to results.
Accordingly, our next effort is aimed at determining the most informative figure (MIF), which will be the one that helps a reader to understand the main idea presented in the paper.
In an academic paper, the abstract can be expected to contain the ideas presented in the entire paper just as figure captions will explain the figure contents.
Therefore, the figure caption that shares the most words with the abstract will more likely belong to the MIF.
The text boxes of captions are estimated using YOLOv2 bounding boxes by calculating the length of vertical direction between bottom of YOLOv2 bounding box and embedded text box corresponding to the caption.

\subsection{Important sentence extraction}
\label{sec:sentence}
To extract the key sentences that express the most important information such as the contributions of the paper, we adapted the informative sentence extraction method proposed by Luhn \cite{luhn1958automatic} that identifies the sentences containing the most important words.
Since these important words must appear repeatedly, their importance can be scored by referring to the frequency of their appearance.
These important sentences are then extracted by referring to the frequency of the important words that are included in the sentence.
Accordingly, we begin by inputting all of the sentences extracted from the PDF file, and then acquire the most informative sentences.

\begin{table}[t]
\def\@captype{table}
  \begin{minipage}[c]{.45\textwidth}
  \tblcaption{Number of annotations in ICCV 2017 dataset.}
   \begin{center}
      \begin{tabular}{p{3.65em}c}
        Class&Number \\
        \hline
        Abstract&621 \\
        Author&621 \\
        Figure&4213 \\
        Title&621 \\
        Table&1995 \\
        \hline
      \end{tabular}
    \end{center}
    \label{tab:annotation}
  \end{minipage}
  \hfill
  \begin{minipage}[c]{.45\textwidth}
 \vspace{8pt}
  \tblcaption{Quantitative result of training on YOLOv2.}
    \begin{center}
      \begin{tabular}{p{3.65em}c}
        Class&IoU \\
        \hline
        Overall&0.52 \\
        Abstract&0.77 \\
        Author&0.5 \\
        Figure&0.56 \\
        Title&0.44 \\
        Table&0.35 \\
        \hline
      \end{tabular}
    \end{center}
    \label{tab:result}
  \end{minipage}
\end{table}

\begin{table}[t]
\def\@captype{table}
\tblcaption{Result of text boxes estimation on papers accepted by CVPR 2018.}
\begin{center}
\begin{tabular}{p{3.65em}ccc}
Class&Completely success&Partially success&Fail\\
\hline
Abstract&483&330&166\\
Author&339&607&33\\
Title&538&404&37\\
\hline
\end{tabular}
\label{tab:textbox}
\end{center}
\end{table}

\section{Experiments}
\label{sec:experiments}
To create a dataset, we collected the 621 of the accepted papers presented at ICCV 2017 and annotated them manually, as shown in Tab.~\ref{tab:annotation}, in order to train YOLOv2 \cite{redmon2017yolo9000}.
For YOLOv2 training, we divided the annotations by the ratio of 8:2 for training and testing, respectively.

\subsection{YOLOv2 training}
We trained YOLOv2 for 10 epochs using the annotated ICCV 2017 dataset.
To evaluate the training quantitatively, we then calculated the intersection over union (IoU) on testing set, as shown in (Tab.~\ref{tab:result}) below.
\begin{equation}
IoU(A,B)=\frac{\mid A\cap B\mid}{\mid A\cup B\mid}
\end{equation}

Here, $A$ and $B$ are regions indicated by bounding boxes.

Additionally, we also evaluate the performance of text boxes extraction, described in Sec.~\ref{sec:visual}, on the 979 papers accepted by CVPR 2018.
We classify the result into three levels, completely success (extracted text boxes exactly much the ground truth), partially success (50 \% of correct text boxes embedded in PDF were extracted or all correct text boxes were extracted, but incorrect one was also obtained), and fail (less than 50 \% of correct text boxes embedded in PDF were extracted) as in Tab.~\ref{tab:textbox}.
The author and title tend to be confused due to the similar appearance, but most papers are classified as completely success or partially success.
Although IoUs of the author and title are much lower than the abstract, their text boxes embedded in PDF can be extracted using YOLOv2 bounding box, which indicates that visual information is useful to obtain textual information.
On the other hand, the extracted text boxes as the abstract often include that of the title of first section, but which is not a problem as only a few words are extracted incorrectly and does not affect the MIF selection described in Sec.~\ref{sec:MIF}.

\subsection{Single-page summary generation}
\begin{figure}[H]
\begin{center}
\includegraphics[width=0.8\linewidth]{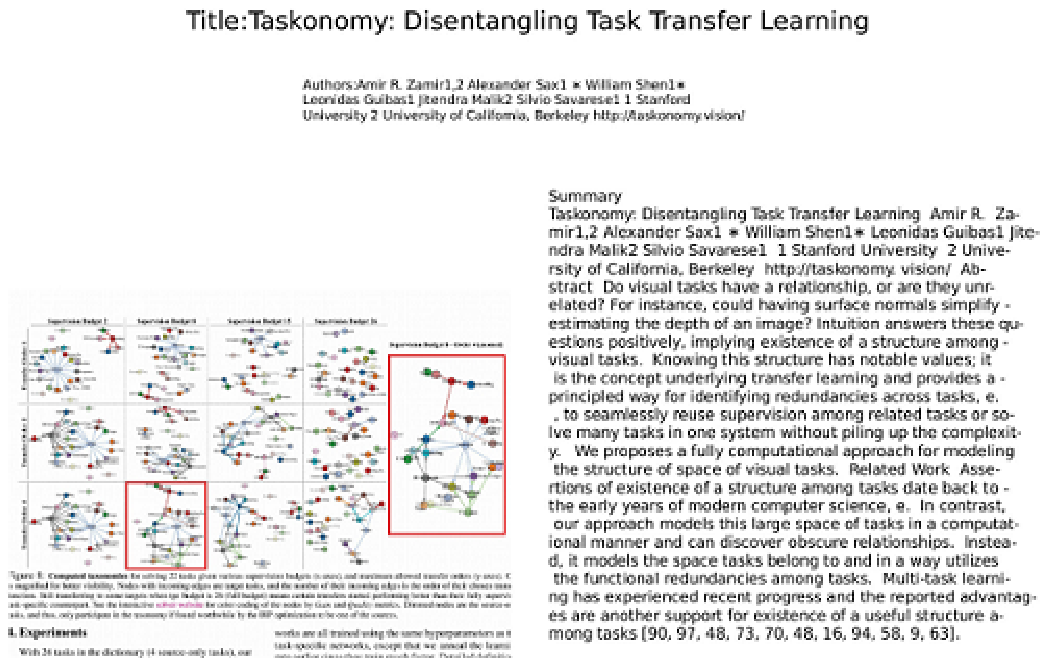}
\caption{Generated summary from \cite{zamir2018taskonomy}}
\label{fig:result1}
\end{center}
\end{figure}

\begin{figure}[H]
\begin{center}
\includegraphics[width=0.8\linewidth]{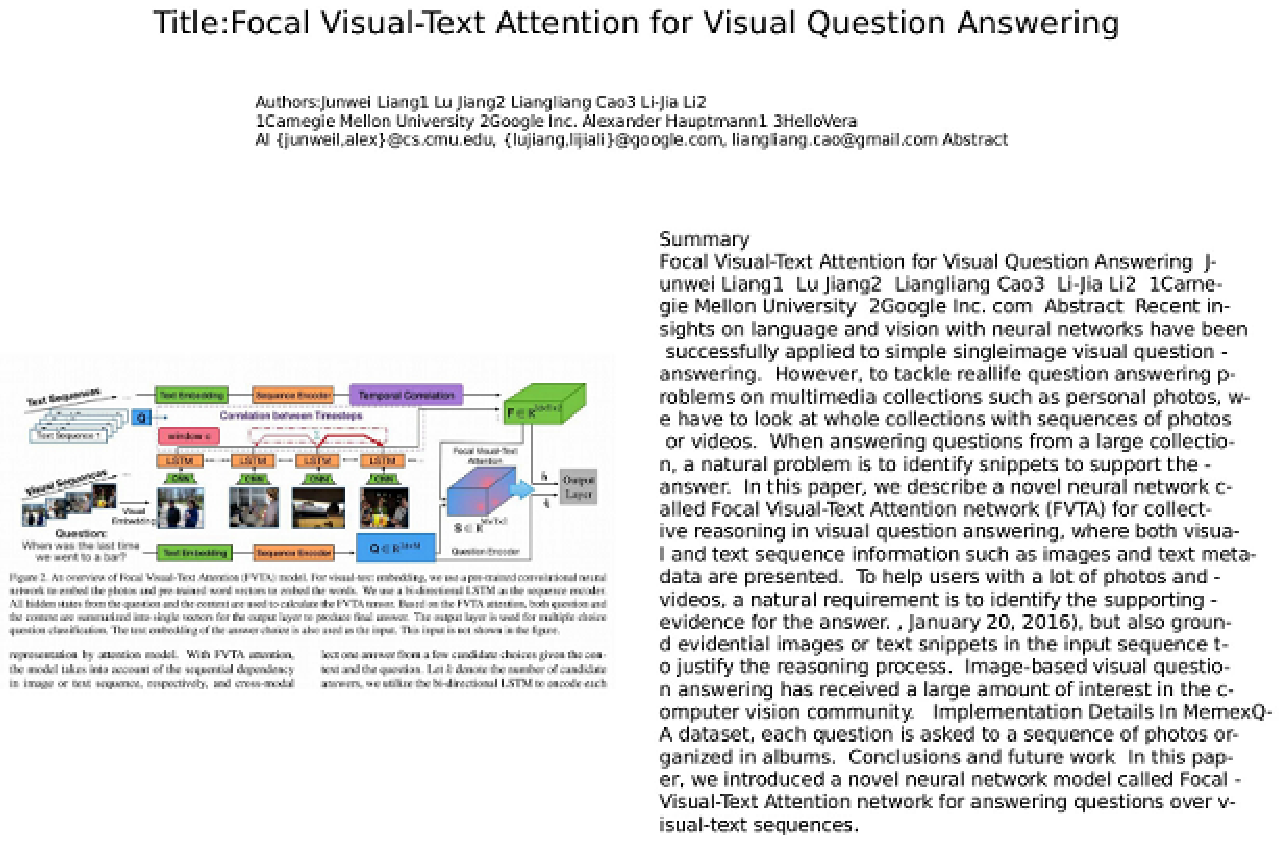}
\caption{Generated summary from \cite{liang2018focal}}
\label{fig:result2}
\end{center}
\end{figure}

The summary of this paper is created using our proposed PSG (Fig.~\ref{fig:sumary_this_paper}).
Since we trained YOLOv2 using the ICCV 2017 dataset, we converted the paper to the format used by ICCV.
Moreover, we also generate the summaries of all papers accepted for CVPR 2018, which will be released.
Some examples of the results are shown in Fig.~\ref{fig:result1} and \ref{fig:result2}.

To analyze the summarization result, we obtain the word frequency of the summary sentences as described in Sec. \ref{sec:sentence} (Tab.~\ref{tab:frequency}).
The result indicates that well-handled topic in CV such as \textit{image} or \textit{learning} is of course included in the summary.
Additionally, the indicators of the topic (e.g. \textit{propose} or \textit{problem}) and the solution (e.g. \textit{approach} or \textit{using}), which help the reader to realize the main topic presented in an academic paper, also appear frequently.
However, some words like \textit{University}, which is the author's information, are not relevant to the main topic presented in an academic paper.
This is not a big problem because we can exclude such a kind of information using extracted text boxes as described in Sec.~\ref{sec:visual}.

\begin{table}[t]
\def\@captype{table}
\tblcaption{Word frequency in the summary sentences (top-20).}
\label{tab:frequency}
\begin{minipage}[c]{.45\textwidth}
\begin{center}
\begin{tabular}{p{3.65em}c}
Word&
The number of occurrences\\
\hline
image&1321\\
University&964\\
images&894\\
Abstract&878\\
network&878\\
learning&853\\
propose&806\\
can&797\\
model&789\\
paper&750\\
\hline
\end{tabular}

\end{center}
\end{minipage}
\hfill
\begin{minipage}[c]{.45\textwidth}
\begin{center}
\begin{tabular}{p{3.65em}c}
Word&
The number of occurrences\\
\hline
deep&621\\
object&603\\
data&560\\
training&554\\
problem&554\\
method&513\\
video&503\\
methods&483\\
using&479\\
approach&448\\
\hline
\end{tabular}

\end{center}
\end{minipage}
\end{table}

\section{Conclusion}
\label{sec:conclusion}
In this paper, we introduced our fully automatic PSG process to overcome the recent rapid increase in the number of published academic papers.
However, our PSG still needs to be improved.
Currently, only one figure is selected as an MIF, but papers often contain multiple figures that represent a variety of information ranging from the method overview to the experimental results, which are designed to help readers obtain a deeper understanding of the subject manner.
Thus, it will be necessary to define the importance of figures on a multi-level scale.
In addition, it will be necessary to improve the generation of summary sentences, as we are currently unable to consider context.
Since citation-based methods \cite{mei2008generating,qazvinian2008scientific,abu2011coherent} cannot be applied to state-of-the-art papers, representing an entire paper is challenging task due to the diversity of the information it contains.
Thus, a novel representation of the main idea presented in the paper is required.

\section*{Acknowledgment}
This work was granted in part by the Strategic Basic Research Program ACCEL of the Japan Science and Technology Agency (JPMJAC1602). We have had the support and encouragement of cvpaper.challenge group.

\bibliography{ref}
\bibliographystyle{spiebib} 

\end{document}